\documentclass[10pt,twocolumn,letterpaper]{article}

\usepackage{iccv}
\usepackage{times}
\usepackage{epsfig}
\usepackage{graphicx}
\usepackage{amsmath}
\usepackage{amssymb}
\usepackage{multirow}
\usepackage[dvipsnames]{xcolor}

\usepackage[breaklinks=true,bookmarks=false]{hyperref}

\iccvfinalcopy 

\def\iccvPaperID{****} 

\begin{document}

\title{MortonNet: Self-Supervised Learning of Local Features in 3D Point Clouds}

\author{Ali Thabet\thanks{indicates equal contribution.} \and
Humam Alwassel\footnotemark[1] \and
Bernard Ghanem \and
King Abdullah University of Science and Technology (KAUST), Saudi Arabia\\
{\tt\small \{ali.thabet,humam.alwassel,bernard.ghanem\}@kaust.edu.sa}
}

\maketitle

\begin{abstract}
We present a self-supervised task on point clouds, in order to learn meaningful point-wise features that encode local structure around each point. Our self-supervised network, named MortonNet, operates directly on unstructured/unordered point clouds. Using a multi-layer RNN, MortonNet predicts the next point in a point sequence created by a popular and fast Space Filling Curve, the Morton-order curve. 
The final RNN state (coined Morton feature) is versatile and can be used in generic 3D tasks on point clouds. 
In fact, we show how Morton features can be used to significantly improve performance (+$3\%$ for $2$ popular semantic segmentation algorithms) in the task of semantic segmentation of point clouds on the challenging and large-scale S3DIS dataset. 
We also show how MortonNet trained on S3DIS transfers well to another large-scale dataset, vKITTI, leading to an improvement over state-of-the-art of $3.8\%$. 
Finally, we use Morton features to train a much simpler and more stable model for part segmentation in ShapeNet. Our results show how our self-supervised task results in features that are useful for 3D segmentation tasks, and generalize well to other datasets.
\vspace{-13pt}
\end{abstract}

\newcommand{\ahat}{\hat{\textbf{a}}}
\newcommand{\av}{\textbf{a}}
\newcommand{\bv}{\textbf{b}}
\newcommand{\cv}{\textbf{c}}
\newcommand{\dv}{\textbf{d}}
\newcommand{\uv}{\textbf{u}}
\newcommand{\vv}{\textbf{v}}
\newcommand{\x}{\textbf{x}}
\newcommand{\X}{\textbf{X}}
\newcommand{\y}{\textbf{y}}
\newcommand{\Y}{\textbf{Y}}
\newcommand{\z}{\textbf{z}}
\newcommand{\w}{\textbf{w}}
\newcommand{\W}{\textbf{W}}
\newcommand{\p}{\textbf{p}}
\newcommand{\q}{\textbf{q}}
\newcommand{\h}{\textbf{h}}
\newcommand{\A}{\textbf{A}}
\newcommand{\B}{\textbf{B}}
\newcommand{\D}{\textbf{D}}
\newcommand{\V}{\textbf{V}}
\newcommand{\U}{\textbf{U}}
\newcommand{\I}{\textbf{I}}
\newcommand{\PX}{\textbf{P}}
\newcommand{\mSigma}{\mathbf{\Sigma}}
\newcommand{\0}{\mathbf{0}}
\newcommand{\1}{\mathbf{1}}

\section{Introduction}
Given their massive success with 2D data, deep learning algorithms are a natural option to solve 3D computer vision problems. The most common form of 3D data comes in the form of unstructured 3D point clouds. However, applying popular 2D based CNNs to point clouds remains an elusive and challenging problem. The main challenge comes from the fact that point clouds are both unstructured and unordered. These properties make typical CNNs unsuitable for point clouds, as they assume a uniformly structured input. To overcome these challenges, researchers have tried to render point clouds into multiple 2D views \cite{mv_su2015multi,mv_shi2015deeppano,mv_guerry2017snapnet}, or voxelize them to apply 3D CNNs \cite{voxel_yi2017large,voxel_riegler2017octnet}. These techniques are either computationally expensive or come with inevitable loss of information \cite{Landrieu_2018_CVPR}. Such drawbacks induce a negative effect on overall performance \cite{pc_huang2018recurrent}. 

A more desirable approach is to operate directly on the point cloud itself. To that extent, PointNet \cite{pc_qi2017pointnet} paved the way to applying deep learning directly to point clouds and served as a gateway for a variety of works in this area. All these works generally follow a two-step process: point-wise computations followed by an aggregation pipeline. The point-wise computations are usually through fully connected or 1D convolutions layers, while the aggregations range from simple global feature pooling to complex RNN architectures. Aggregation is usually done over a well-defined and structured neighborhood, \eg spatial columns or 3D grids. However, these neighborhoods are not generally nor necessarily related to the structure of the point and its 3D surrounding area. 

\begin{figure}[t]
    \centering
    \includegraphics[width=0.98\linewidth]{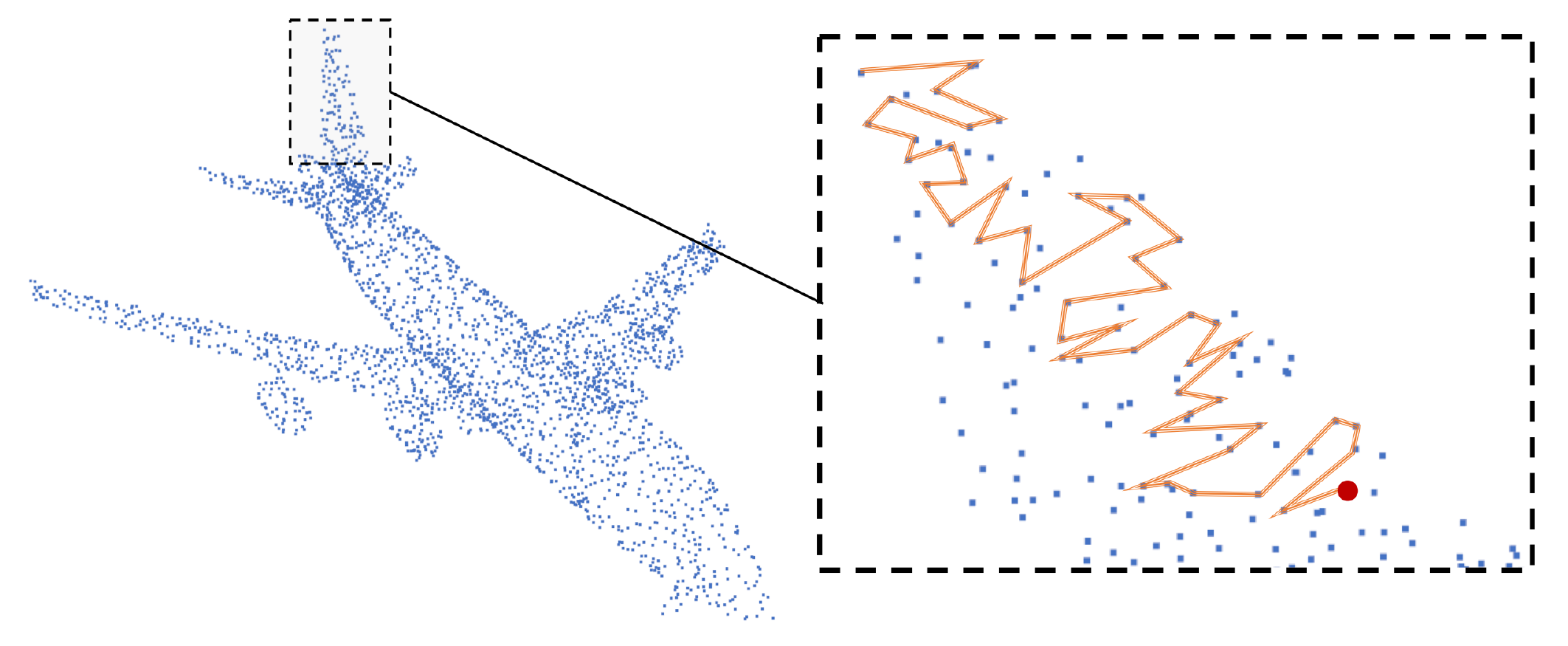}
    \caption{\small\textbf{Self-Supervision of 3D Point Clouds.} Applying deep learning techniques on point clouds is challenging due to their unstructured and unordered nature. We propose to learn a point-wise feature representation by leveraging space filling curves, namely the Morton- or Z-order (refer to point sequence in orange in the plane's tail). Given this ordering, we setup a self-supervised task that aims to predict a point (the red point) from the sequence points (the orange points). Interestingly, features learned from this self-supervision can be used to improve performance in popular 3D tasks such as semantic segmentation.    }
    \label{fig:pull}
\end{figure}

We propose learning a new type of point-wise feature by using points and a structured neighborhood around them. We define this neighborhood based on a predefined meaningful geometric ordering for 3D points, namely the well-known Morton-order (also known as Z-order) Space Filling Curve (SFC). Given this ordering, a self-supervised task of point prediction can be indirectly exploited to extract local point-wise features, which can be later used in common 3D tasks including semantic or part segmentation (refer to \ref{fig:pull}).  
To the best of our knowledge, this is the first time self-supervised learning is exploited with 3D point clouds.


\vspace{8pt}\noindent \textbf{Contributions:} We summarize our contributions as three-fold. \textbf{(1)} We propose a self-supervised task and accompanying network to learn point-wise features in 3D point clouds; we denote this network as MortonNet, and consequently, the features it learns as Morton features. To the best of our knowledge, we are the first to apply self-supervision to 3D point clouds. 
\textbf{(2)} Since Morton features are learned in a self-supervised manner, we show how they can be easily incorporated in popular 3D tasks. Specifically, we show how they can be used in semantic segmentation pipelines to substantially increase their performance by over $3\%$ on the challenging and dense dataset S3DIS \cite{s3dis}.
\textbf{(3)} We also demonstrate how another advantage of self-supervised learning is inherited, namely model generalization to different datasets. Specifically, we show how MortonNet features trained on S3DIS successfully transfer to two other datasets: vKITTI \cite{pc_engelmann2018}, where state-of-the art performance in  semantic segmentation is improved by $3.8\%$,  and ShapeNet \cite{shapenet}, where training is significantly simplified for   the task of part segmentation while maintaining performance comparable to current state-of-the-art algorithms.


In what follows, we first review current 3D deep learning algorithms in Section \ref{sec:related}; we talk briefly about self-supervision in non 3D applications as well as review Morton-order curves. We present the self-supervised task on Morton-ordered point sequences in Section \ref{sec:method}. In Section \ref{sec:exp}, we presents our experiments and results, and finally, we draw conclusions in Section \ref{sec:conc}.

\section{Related Work}
\label{sec:related}
\subsection{Deep Learning for 3D Data}
In this section, we give an overview of some related works that touch upon different facets of 3D data analysis. We focus here on 3D deep learning algorithms. The principal separating factor between these algorithms is how they represent the input data. More specifically, 3D data can be represented as voxelized volumes, 2D multi-views, or unstructured point clouds. Voxelized methods convert 3D data into regular volumetric occupancy grids (voxels), which form a structured volume well-suited for 3D CNNs. Earlier methods trained end-to-end 3D CNNs for several 3D tasks \cite{voxel_dai2017scannet,voxel_huang2016point,voxel_maturana2015voxnet,voxel_mv_qi2016volumetric}. Voxelized representations are constrained by their resolution due to the volumetric sparse nature of 3D data and the expensive 3D convolutions they use. Several works try to address this issues \cite{voxel_engelcke2017vote3deep,voxel_graham2015sparse,voxel_graham2014spatially,voxel_riegler2017octnet,voxel_tchapmi2017segcloud}. Multi-view methods project 3D data into multiple 2D views and use popular 2D CNNs to process them. Qi \etal \cite{voxel_mv_qi2016volumetric} present an extensive study of multi-view methods and their performance related to voxelized approaches.

A more lucrative approach is to operate directly on unordered point clouds. To that extent, PointNet \cite{pc_qi2017pointnet} pioneered the way for deep learning methods to be applied to unordered point clouds. The authors show the benefit of using this data modality as opposed to either voxelized or multi-view methods. PointNet takes point cloud chunks, computes point features, and aggregates those features with an order-invariant operation like max pooling. Pooled features are used to either label the whole chunk (classification) or each point (segmentation). Inspired by this line of work, new research presents better ways to compute and aggregate point features by either looking at more local context \cite{pc_qi2017pointnet++, pc_engelmann2018} or using more complex RNN based methods \cite{pc_huang2018recurrent,pc_ye20183d}.

All these works try to incorporate local information for each point, by pooling from some structured neighborhood. The local representation of these features highly depends on the type of neighborhood used. Wang \etal \cite{dgcnn} propose a graph CNN algorithm for point cloud semantic segmentation where local neighborhoods are dynamically changed based on nearest neighbors at each layer, while Tatarchenko \etal  propose learning local features using Tangential Convolutions \cite{pc_tatarchenko2018tangent}. Both \cite{dgcnn} and \cite{pc_tatarchenko2018tangent} define more flexible local environments for points. However, as with other point cloud methods, their task is fully supervised end-to-end. In contrast, we propose learning point features based on local structure, but we do it in a self-supervised manner. Our approach creates point neighborhoods defined by a meaningful geometric ordering (Morton- or Z-ordered curves). Our extensive experiments show that these features can significantly boost the performance of 3D learning pipelines and generalize well to different datasets.

\subsection{Self-Supervision}
Self-supervision is a form of unsupervised learning, where supervision originates from the data itself. The principal idea is to set a task that only depends on the data available. This task serves as a proxy to learn semantic representations.
With images, self-supervision has been applied to learn relative patch locations from the same image \cite{ss_doersch2015unsupervised}, learn colorization \cite{ss_zhang2016colorful}, design examplar CNNs \cite{ss_dosovitskiy2016discriminative}, and learn image rotations \cite{gidaris2018unsupervised}. In video, examples of self-supervision work are learning the next frame in a video sequence \cite{ss_vondrick2015anticipating} and tracking by colorization \cite{ss_vondrick2018tracking}. For 3D data, few works use unsupervised methods to learn volumetric representations of 3D data \cite{ss_tulsiani2017learning,ss_tatarchenko2017octree}. And, self-supervision has not been successfully transferred to deal with 3D point clouds yet. However, self-supervised algorithms do exist for graphs, where the data shares the unstructured nature inherent to point clouds, and several of its challenges \cite{ss_hamilton2017inductive, ss_zhang2017weisfeiler}.

\subsection{Space Filling Curves} \label{subsection:space_filling_curves}
Space Filling Curves (SFC) are mappings from multidimensional data to one dimension. They generally have good locality-preserving behavior (\ie direct Euclidean  neighbors in 1D tend to be similar to those in 3D). In the context of point clouds, SFCs are popular for nearest neighbor algorithms \cite{sf_connor2010fast}. A common choice of SFC is the Morton-order, usually referred to as the Z-order \cite{sf_morton1966computer}. Z-order curves are popular due to their low computational complexity. Other SFC choices include Hilbert, Peano, Sierpinski, among others. We direct the reader to the extensive review by Hans Sagan \cite{sf_sagan2012space} of different types of SFCs. We use SFCs to define a geometric ordering suitable for our self-supervised task. We choose the Z-order specifically because of its good balance between locality preservation and computational complexity. Refer to the \textbf{supplementary material} for details about Morton-order and its implementation.

In this paper, we present a new method to compute meaningful point features for unstructured point clouds. We compute these features through a self-supervised task that does not require semantic labels. Contrary to previous work, our features can represent the local structure around each point. We show how these features can be incorporated to perform common 3D tasks, specifically semantic and part segmentation, and how  they are generalizable to other datasets.  

\begin{figure*}[!htb]
    \centering
    \includegraphics[width=\linewidth]{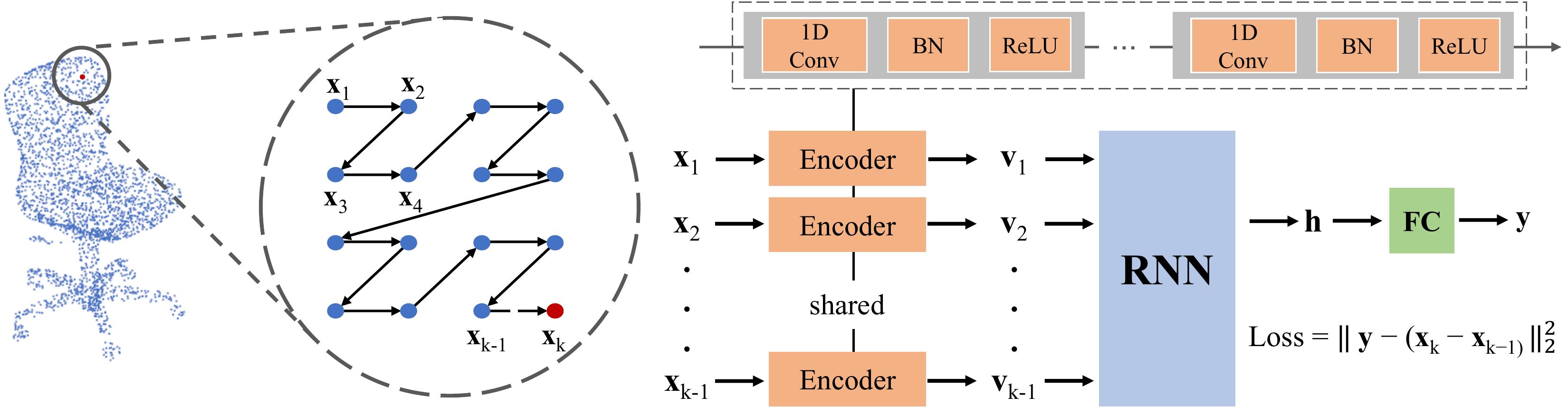}
    \caption{\textbf{Learning Morton Features.} MortonNet aims to learn local features of unstructured 3D point clouds. Our model is trained in a self-supervised fashion to predict the displacement to the next point in a Z-ordered sequence. Given an unstructured point cloud, we first generate multiple Z-ordered sequences that end at each point (\eg the red point $\x_k$). We encode each point $\x_i$ in the sequence (except $\x_k$) into a higher dimensional vector $\vv_i$ with a series of 1D convolutions, batch normalization, and ReLU. The sequence of vectors $(\vv_1, \vv_2, \dots, \vv_{k-1})$ are then fed into a multi-layer RNN, whose final state $\h$ is transformed by a fully connected layer to produce the final displacement prediction $\y$. We regress $\y$ to the ground truth displacement $(\x_k - \x_{k-1})$ for each Z-ordered sequence.}
    \label{fig:pipeline}
\end{figure*}

\section{Methodology}
\label{sec:method}
To represent 3D points in a way that captures meaningful detail for generic point-based 3D tasks (\eg part and semantic segmentation), we aim to learn local  features from  unstructured point clouds. In this section, we formulate a self-supervised proxy task to learn such  features (refer to \ref{fig:pipeline} for an overview of the model used and the learning setup). We highlight the process to generate the self-supervised training data from any point cloud, and apply it to the large scale and dense S3DIS dataset \cite{s3dis}. Finally, we detail our proposed model architecture to solve the self-supervised task and give its implementation details.

\subsection{Predict-the-Next-Point Self-Supervised Task} 
\label{subsection:ss_task}
Given a $k$-long Z-ordered sequence ($\x_1$, $\x_2$, \dots, $\x_k$), the task is to predict $\x_k$ from the previous $(k-1)$ points in the sequence. To stabilize the learning process, we consider the equivalent task where we predict the displacement $\x_k - \x_{k-1}$ given the $(k-1)$-long subsequence ($\x_1$, $\x_2$, \dots, $\x_{k-1}$). The Z-order gives us a stable structure to learn from any unstructured point cloud. We turn to self supervision to learn this task since it does not require annotated data. Next, we detail how Z-ordered sequences can be generated from an unstructured point cloud.

\subsection{Z-ordered Sequence Generation}
\label{subsection:sequence_generation}
Let $\mathcal{S}_{r}(\x)$ be the set of points that lie in a neighborhood of radius $r$ around the point $\x$. We refer to $\mathcal{S}_{r}(\x)$ as the support of $\x$. Given a point $\x$ in our 3D point cloud, we generate Z-ordered sequences that end at $\x$ in the following way: we order all points in $\mathcal{S}_{r}(\x)$ according to their Z-order (refer to Subsection \ref{subsection:space_filling_curves}), then we randomly sample $(k-1)$ points from $\mathcal{S}_{r}(\x)$ that have a smaller Z-order compared to that of $\x$. Finally, we construct a Z-ordered sequence of length $k$ that ends at $\x$, \ie the $(k-1)$ sampled points plus $\x$. We generate $m$ different Z-ordered sequences in the same manner for every point $\x$ in the point cloud. 
Our sequence generation procedure guarantees that we cover each point with multiple sequences, allowing us to capture the different aspects of local structure in the point cloud.

\subsection{Proposed MortonNet Architecture}
The input to our MortonNet model is a ($k-1$)-long Z-ordered sequence of 3D points, ($\x_1$, $\x_2$, \dots, $\x_{k-1}$), and the expected output is the displacement to the next point, \ie $(\x_k - \x_{k-1})$. We encode each 3D point $\x_i$ into a higher dimensional vector $\vv_i$ using a series of spatial encoding layers, each of which consists of a set of 1D convolutions followed by batch normalization and a ReLU activation function. Then, we feed the high dimensional vector sequence ($\vv_1$, $\vv_2$, \dots, $\vv_{k-1}$) to a multi-layer RNN. Finally, we transform the last RNN hidden state using a fully connected layer to produce a 3D final output $\y$, \ie an estimate of the spatial displacement needed to reach the next point in the sequence. 
Figure \ref{fig:pipeline} gives an overview of the MortonNet  architecture.

\subsection{Implementation Details}
\noindent\textbf{Sequence Generation.} 
We select the minimum radius $r$ for which we have at least $2 \times k$ points in the support set $\mathcal{S}_{r}(\x)$. We experiment with different sequence length values, namely $k = 20$, $60$, and $100$. We note that $k$ defines the number of points encoding the local structure around a point. It is therefore possible to select shorter sequences for areas of simpler structure like planes. For simplicity however, we set  all sequences to the same length. Finally, we empirically set the number of sequences per point to $m = 5$. 

\vspace{8pt}\noindent\textbf{Architecture.}
\label{subsection:architecture}
We use between $2$ and $4$ encoding layers with $64$ 1D-convolution kernels in each. We choose a GRU architecture for the RNN with $3$ or $4$ layers and a $200$ or $300$ hidden state size. After learning to predict the next point in a Z-ordered sequence, we extract features from the last RNN hidden state $\h$ of each sequence. Since we have $m$ sequences that end at the same point, the Morton feature we assign to that point is the max pooling of the $m$ states $\h$.

\vspace{8pt}\noindent\textbf{Training Procedure.} 
Since local geometry is translationally invariant (\eg the local structure of a chair arm does not change by moving the chair itself in space), we make each Z-ordered sequence translationally invariant by subtracting the coordinate of the first point from each point in the sequence. We regress the model output $\y$ to the ground truth displacement $(\x_k - \x_{k-1})$ using the mean square error loss. We use an Adam optimizer and train for $40$ epochs. We set the initial learning rate to $10^{-3}$ and decrease the learning rate with a decay factor of $0.9$ when the validation loss does not improve for $2$ epochs. We pick the final model at the epoch with the best validation loss. 


\section{Experiments}
\label{sec:exp}
We train our model on the Predict-the-Next-Point task (Section \ref{subsection:ss_task}) on the Stanford Large-Scale 3D Indoor Spaces (S3DIS) dataset \cite{s3dis}. We choose S3DIS, since it is a large-scale dataset with high density point clouds and a large number of object instances. These characteristics translate into a large variety of Z-ordered sequences for our model.  We incorporate the learned Morton features (Section \ref{subsection:architecture} 
) into the popular tasks of point cloud semantic and part segmentation. We also show the power of Morton features by demonstrating how the learned features transfer between datasets. We present how Morton feature-based 3D classifiers can significantly enhance the performance of methods using S3DIS, achieve state-of-the-art semantic segmentation results on vKITTI \cite{pc_engelmann2018}, and simplify the part segmentation task on ShapeNet \cite{shapenet}. 

We organize this section as follows. We present the  datasets we use to train our self-supervised model and those for semantic and part segmentation. Then, we present the experimental setting and results for learning local features in S3DIS. Finally, we present experiments that show the benefits of Morton features in semantic segmentation of S3DIS and their transferablity to other datasets and tasks.

\subsection{Dataset Details}
\noindent\textbf{S3DIS \cite{s3dis}.} 
The Stanford Large-Scale 3D Indoor Spaces (S3DIS) dataset consists of indoor 3D point clouds of $6$ large areas (complete floors) from several buildings. The indoor scans cover an area of $6020$ meters and contain over $215$ million points. The data is semantically separated into $272$ rooms and annotated with $12$ semantic elements, $7$ of which are related to structure such as \emph{ceiling} and \emph{door}, while the rest
relate to common furniture items such as \emph{chair} and \emph{bookcase}. There is an additional label for clutter. S3DIS is typically used for indoor semantic segmentation. 

\vspace{8pt}\noindent\textbf{vKITTI \cite{pc_engelmann2018}.} Virtual-KITTI (vKITTI) is a synthetic large outdoor dataset with $13$ semantic classes from urban scenes. vKITTI imitates data from the real-world KITTI dataset. It contains data from $5$ different simulated worlds, resulting in $50$ high resolution scenes. This dataset is used for a variety of tasks, where the most common one in regards to point clouds is semantic segmentation.

\vspace{8pt}\noindent\textbf{ShapeNet \cite{shapenet}.}
This synthetic dataset consists of about $32000$ 3D CAD models belonging to $16$ shape categories from the original ShapeNetCore 3D data repository \cite{shapenetcore}. Each point in a 3D model is annotated with a part label (\eg a plane is segmented into body, wing, engine, and tail parts). We consider the subset used for the ShapeNet part segmentation challenge, which contains $17775$ models with $50$ parts in total. Each model is normalized into the 3D cube $[-1,1]^3$ and contains a maximum of $3000$ points. We follow the data split of \cite{pc_qi2017pointnet++}, where  $90\%$ of each class's point clouds are used for training and the rest for testing. This dataset is typically used for part segmentation.

\subsection{Learning Morton Features} \label{subsection:ss_experiment}
In this section, we investigate how local structural features can be learned by predicting the next point in a Z-ordered sequence. Refer to Section \ref{subsection:ss_task} for more details about the task formulation. 

\vspace{8pt}\noindent\textbf{Experimental Settings.} 
We study the performance of our MortonNet model on this self-supervised task on the S3DIS dataset. We discard the semantic label information in S3DIS and only use the unlabelled point clouds. 
We train MortonNet on a subset of the original data. In particular, we select $8$ rooms from the training subset and $2$ from the testing subset. This constitutes less than $4\%$ of all rooms. We first generate the $k$-long Z-ordered sequences for all the points in the rooms, and later subsample $40\%$ of the training sequences for computational efficiency.




\vspace{8pt}\noindent\textbf{Baseline Methods.} To demonstrate the effectiveness of learning from Morton-ordered sequences, we train baseline models on the same predicting-the-next-point task from sequences ordered using different mechanisms. The \textit{Random Sequence Baseline} is trained on sequences of points that are randomly ordered. We choose this model for reference, so as to show that learning this self-supervised task is more meaningful when the sequence is ordered. The \textit{Coordinate-Ordered Sequence Baselines} train three models on point sequences ordered according to their $x$-, $y$-, and $z$-coordinates. We choose these models to show that Morton-order is a better ordering choice. 

\vspace{8pt}\noindent\textbf{Evaluation Metric.}
To measure the accuracy of our model and the baselines in predicting-the-next-point, we consider a prediction to be correct if it is within a ball of radius $\rho$ from the correct next point in the sequence. In other words, if the ground truth displacement is $(\x_k - \x_{k-1})$, and the  the model predicts a displacement value $\y$, then the prediction is correct if and only if $\| \y - (\x_k - \x_{k-1}) \|_2 \le \rho$. We set $\rho = 0.02$ in our experiments and report the performance as the average accuracy on the testing sequences.  

\vspace{8pt}\noindent\textbf{Results.}
Table \ref{table:ss_results} reports the test accuracy of our model against the baselines on S3DIS. MortonNet is consistently superior to both baselines. This shows the importance of the Z-order in learning local point features from unstructured point clouds. Additionally, we show the effect of the parameter $k$ (the sequence length) on the accuracy. As $k$ increases, MortonNet becomes more accurate in predicting the next point. We attribute this correlation between $k$ and accuracy to the fact that with longer Z-ordered sequences, MortonNet has access to a bigger support set $\mathcal{S}_r(\x)$ and thus can encode more complex local structures around each point. However, while increasing $k$ beyond $100$ would result in a more accurate model, the increase in performance comes at the cost of making the model computationally expensive, since RNNs process data sequentially. Thus, we choose $k=100$ for all remaining experiments. 
\begin{table}[t!]
    \small
    \tabcolsep=0.5cm
    \vspace{6pt}
    \begin{center}
        \caption{\textbf{MortonNet vs. Baselines}. Test accuracy comparison between MortonNet and four baselines (random ordering and three coordinate-wise orderings) on the predict-the-next-point task. We show the results for different sequence lengths $k$. All baselines use $k=100$. MortonNet with $k=100$ gives the best results.} \label{table:ss_results}
        \vspace{8pt}
        \begin{tabular}{l | c }
            \hline
            
            \hline
            Baseline Method & Test Accuracy \\
            \hline
            
            \hline
            Random Sequence                 &  $36.6$ \\
            x-coordinate Ordering           &  $37.6$ \\
            y-coordinate Ordering           &  $36.9$ \\
            z-coordinate Ordering           &  $37.9$ \\
            \textbf{MortonNet} ($k=20$)     &  $34.0$ \\
            \textbf{MortonNet} ($k=60$)     &  $46.6$ \\
            \textbf{MortonNet} ($k=100$)    &  $\mathbf{84.0}$ \\
            
            \hline            
            
            \hline
        \end{tabular}    
    \end{center}
\end{table}



\begin{table*}[t!]
    \small
    \tabcolsep=0.075cm
    \vspace{6pt}
    \begin{center}
        \caption{\textbf{Semantic Segmentation on S3DIS}. Results of applying Morton features to the S3DIS semantic segmentation task for test Area $5$. The first sub-table shows top performing algorithms that report results on Area $5$. The middle sub-table shows results using PointNet and PointNet with Morton features; * highlights the best performer between both methods. Finally, the last sub-table compares RSNet with and without Morton features (* also highlights best performer between the $2$). Performance is measured by mean intersection of union (mIoU) and mean accuracy (mAcc) across classes. We see clear improvement in mIoU when adding Morton features to both PointNet and RSNet. We also note RSNet + Morton beats current state-of-the-art for Area $5$. Bold numbers represent top performance in mIoU and mAcc.}\label{table:s3dis_seg}
        \vspace{8pt}
        \begin{tabular}{l | c | c | c c c c c c c c c c c c c}
            \hline
            
            \hline
            
            Method & mIoU & mAcc & ceiling & floor & wall & beam & column & window & door & table & chair & sofa & bookcase & board & clutter \\
                                   
            \hline
            
            \hline
            
            SEGCloud \cite{voxel_tchapmi2017segcloud} & $48.9$ & $57.4$ & $90.1$ & $96.0$ & $69.9$ & $0.0$ & $18.4$ & $38.3$ & $23.1$ & $70.4$ & $75.9$ & $40.9$ & $58.4$ & $13.0$ & $41.6$ \\
            DGCNN \cite{dgcnn} & $51.5$ & $ 59.8 $ & $93.0$ & $97.4$ & $77.7$ & $0.0$ & $12.2$ & $47.8$ & $39.8$ & $67.4$ & $72.4$ & $23.2$ & $52.3$ & $39.8$ & $46.6$ \\
            Tangent Convs \cite{pc_tatarchenko2018tangent} & $52.8$ & $62.2$ & $ - $ & $ - $ & $ - $ & $ - $ & $ - $ & $ - $ & $ - $ & $ - $ & $ - $ & $ - $ & $ - $ & $ - $ & $ - $ \\
            3P-RNN \cite{pc_ye20183d}                            & $53.4$ & $\mathbf{71.3}$ & $95.2$ & $98.6$ & $77.4$ & $0.80$ & $9.83$ & $52.7$ & $27.9$ & $78.3$ & $76.8$ & $27.4$ & $58.6$ & $39.1$ & $51.0$ \\
            SPG \cite{Landrieu_2018_CVPR}                   & $54.7$ & $61.7$ & $91.5$ & $97.9$ & $75.9$ & $0.0$ & $14.2$ & $51.3$ & $52.3$ & $77.4$ & $86.4$ & $40.4$ & $65.5$ & $7.23$ & $50.7$ \\
            

            \hline
            \hline
            
            PN \cite{pc_qi2017pointnet}                           & $41.1$ & $49.0$ & $88.8$ & $97.3$ & $69.8$ & $0.05$ & $3.92$ & $49.3$* & $10.8$ & $58.9$ & $52.6$ & $5.8$* & $40.3$ & $26.3$ & $32.2$ \\
            PN + \textbf{MortonNet}                          & $44.4$* & $63.1$* & $90.3$* & $97.9$* & $74.6$* & $0.11$* & $9.55$* & $47.4$ & $17.7$* & $64.2$* & $57.9$* & $1.42$ & $43.2$* & $33.5$* & $40.1$* \\ 
            
            \hline
            \hline
            
            RSNet \cite{pc_huang2018recurrent}               & $51.9$ & $59.4$ & $93.3$* & $98.4$ & $79.2$ & $0.0$ & $15.8$ & $45.4$ & $50.1$* & $67.9$ & $65.5$ & $52.5$* & $22.5$ & $41.0$* & $43.6$ \\
            RSNet + \textbf{MortonNet}                          & $\mathbf{55.0}$* & $61.2$* & $92.5$ & $98.5$* & $81.4$* & $0.0$ & $24.2$* & $47.6$* & $50.0$ & $70.8$* & $76.7$* & $31.6$ & $58.2$* & $38.5$ & $45.1$* \\ 
            
            \hline
        \end{tabular}    
    \end{center}
\end{table*}

\subsection{Morton Features for 3D Tasks} \label{subsection:morton_for_3d_tasks}
We now use the trained MortonNet in three different experiments: semantic segmentation on S3DIS, semantic segmentation on vKITTI, and part segmentation on ShapeNet. Each experiment shows a different way to use Morton features in 3D vision pipelines. Since we can compute features for every point, in theory we could incorporate them in any 3D algorithm. We show how to do this with different algorithm choices for each experiment. 
Our aim is to demonstrate the power of the Morton features we learned by  showing that: (1) they are effective for 3D point-based part and semantic segmentation and (2) they generalize well across different different datasets. 

\vspace{8pt}\noindent\textbf{Experimental Settings.}
We use our MortonNet model (with $k=100$) trained on S3DIS to extract features for semantic segmentation in S3DIS and vKITTI, and for part segmentation in ShapeNet. Using the procedure detailed in Section \ref{subsection:sequence_generation} to extract features, we generate Z-ordered point sequences for all the datasets, pass these sequences to MortonNet, and aggregate the hidden state of the RNN. In all cases, we use the same MortonNet pre-trained on the unlabelled point clouds of S3DIS and we do not finetune on the other datasets.

\vspace{8pt}\noindent\textbf{Evaluation Metrics.} 
We measure performance in all three experiments using the mean point intersection over union (mIoU) across classes, which is the preferred metric for both part and semantic segmentation. We also provide the mean accuracy (mAcc) metric for the semantic segmentation experiments on S3DIS and vKITTI.

\vspace{8pt}\noindent\textbf{Experiment (1): Semantic Segmentation on S3DIS.} Here, we use MortonNet features on the same dataset it was self-supervised on (S3DIS) to show how it can be used to significantly improve the performance of two distinct semantic segmentation algorithms. In this task, we need to classify each point in a room into one of the $13$ semantic indoor labels. 
For the choice of segmentation algorithms, we select the classical PointNet \cite{pc_qi2017pointnet} and the more recent RSNet \cite{pc_huang2018recurrent}, both with readily available code. Note that although the authors of SPG \cite{Landrieu_2018_CVPR} do provide code, training their networks is difficult, since it involves computing hand-crafted features in a non-end-to-end fashion; we consider this an overkill for our purposes. Using the same data processing procedure suggested in \cite{pc_huang2018recurrent}, we spatially divide each room in S3DIS into blocks of 1m$\times$1m along the horizontal direction, and sample $4096$ points from each block. We also use the train/test splits used in \cite{pc_huang2018recurrent}, where we train each segmentation method on Areas $1,2,3,4,$ and $6$, and test on Area $5$. For PointNet, we concatenate our pre-computed Morton features to those extracted in PointNet before the point classifier. 
In RSNet, we concatenate our pre-computed Morton features with their point features before they are fed to their RNN layers. Here, our Morton feature can be seen as an additional local feature that both PointNet and RSNet can use to classify points. For both algorithms, we follow the same training schedule proposed by the original papers. 
Refer to the \textbf{supplementary material} for full implementation details. 

\begin{figure}[h]
    \centering
    \includegraphics[width=0.99\linewidth]{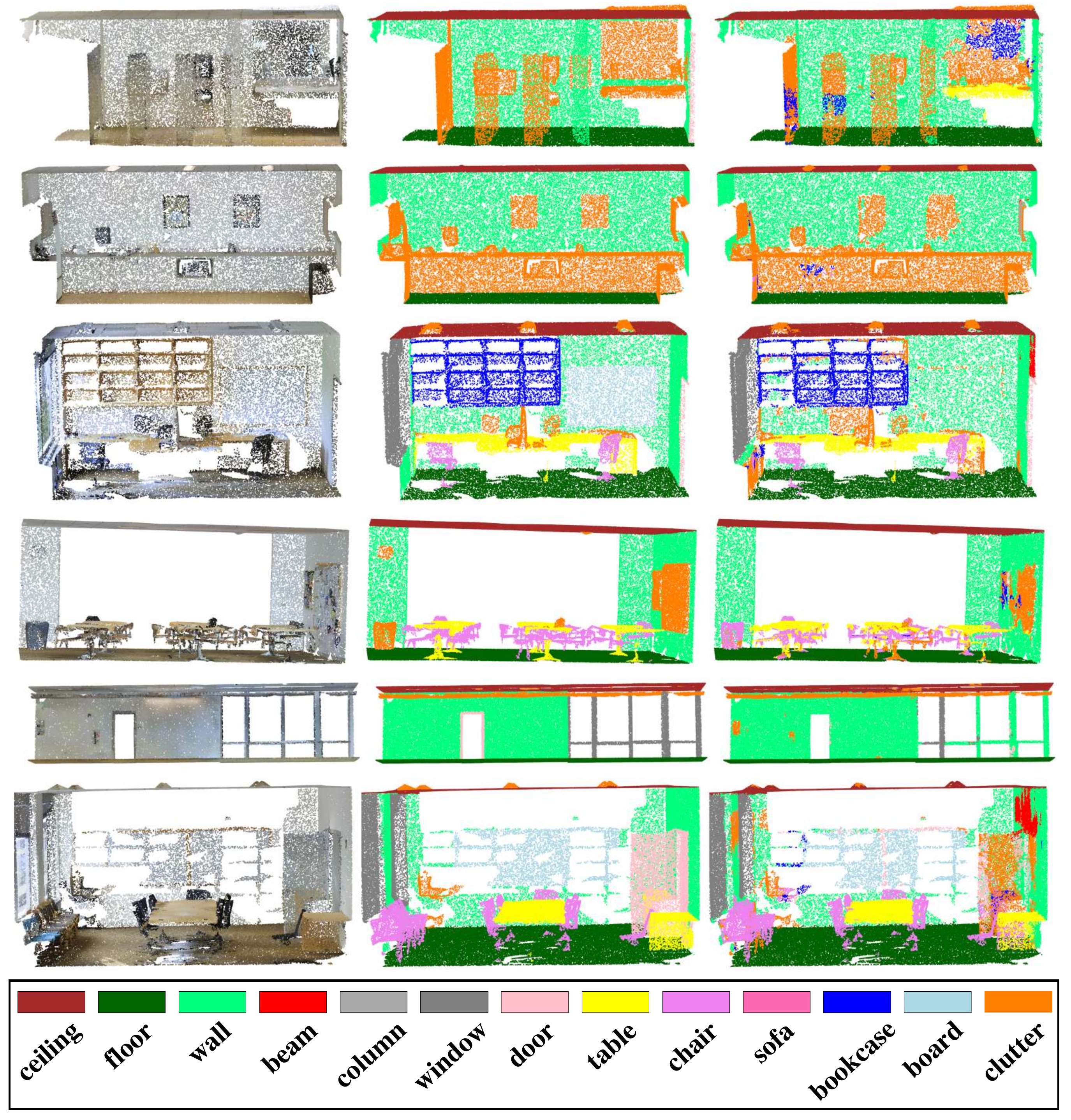}
    \caption{\small\textbf{Semantic segmentation qualitative results on S3DIS} \cite{s3dis}. (\emph{Left}): raw point clouds. (\emph{Middle}): ground truth segmentation. (\emph{Right}): results of PointNet + MortonNet. Better viewed in color and zoomed. Our method performs well with difficult classes like bookcase, clutter, and table.}
    \label{fig:qualitative_s3dis}
\end{figure}

Table \ref{table:s3dis_seg} presents our results for semantic segmentation on S3DIS, along with top performing segmentation methods including the current state-of-the-art, for reference and comparison. We note how augmenting both PointNet and RSNet with Morton features boosts their performance by over $3$ points in mIoU; this is a significant increase given that the current state-of-the-art in Area 5 \cite{Landrieu_2018_CVPR} outperforms the second best \cite{pc_ye20183d} by less than $1$ point in mIoU. In fact, adding Morton features to RSNet outperforms state-of-the-art given by SPG  \cite{Landrieu_2018_CVPR}. These results show that training MortonNet with S3DIS translates to meaningful information about local point structure, which can be used to significantly improve semantic segmentation performance. We also provide qualitative results in Figure \ref{fig:qualitative_s3dis}. Since Morton features are generic point features, they could be added to either of the algorithms in Table \ref{table:s3dis_seg}; we show more details in the \textbf{supplementary material}.

\begin{table}[h]
    \small
    \tabcolsep=0.13cm
    \vspace{6pt}
    \begin{center}
        \caption{\textbf{Semantic Segmentation on vKITTI}. Results of applying Morton features to the vKITTI semantic segmentation task. Performance is measured by overall point accuracy (OA), mean intersection of union (mIoU) and mean accuracy (mAcc) across classes. Note how adding Morton features to PointNet significantly increases its performance and beats current state-of-the-art. Bold numbers represent top performance in mIoU, OA, and mAcc.}\label{table:vkitti_seg}
        \vspace{8pt}
        \begin{tabular}{l | c | c | c}
            \hline
            
            \hline
            
            Method & OA & mIoU & mAcc \\
                                   
            \hline
            
            \hline
            
            G + RCU \cite{pc_engelmann2018} & $80.6$ & $36.2$ & $49.7$ \\
            3P-RNN \cite{pc_ye20183d} & $87.8$ & $41.6$ & $54.1$ \\

            \hline
            \hline
            
            PN \cite{pc_qi2017pointnet} & $79.7$ & $34.4$ & $47.0$\\
            PN + \textbf{MortonNet} & $\mathbf{91.8}$ & $\mathbf{45.4}$ & $\mathbf{63.5}$ \\ 
            \hline
            
            \hline
        \end{tabular}    
    \end{center}
\end{table}
\begin{table*}[t!]
    \small
    \tabcolsep=0.1cm
    \vspace{6pt}
    \begin{center}
        \caption{\textbf{Part Segmentation on ShapeNet}. Performance is measured by mean Intersection of Union (mIoU). Our model uses a very simple classifier on top of Morton features learned using S3DIS. While not an end-to-end approach, our simple architecture achieves comparable results and outperforms state-of-the-art in $7$ out of the $16$ shape categories. Bold numbers represent top performance in mIoU.}\label{table:shapenet_seg}
        \vspace{8pt}
        \begin{tabular}{l | c | c c c c c c c c c c c c c c c c }
            \hline
            
            \hline
            
            \multirow{2}{*}{Method} & \multirow{2}{*}{mean} & \multirow{2}{*}{aero} & \multirow{2}{*}{bag} & \multirow{2}{*}{cap} & \multirow{2}{*}{car} & \multirow{2}{*}{chair} & ear & \multirow{2}{*}{guitar} & \multirow{2}{*}{knife} & \multirow{2}{*}{lamp} & \multirow{2}{*}{laptop} & \multirow{2}{*}{motor} & \multirow{2}{*}{mug} & \multirow{2}{*}{pistol} & \multirow{2}{*}{rocket} & skate & \multirow{2}{*}{table}\\
            & & & & & & & phone & & & & & & & & & board & \\
            
            \hline
            
            \hline
            
            Yi \cite{shapenet} & $81.4$ & $81.0$ & $78.4$ & $77.7$ & $75.7$ & $87.9$ & $61.9$ & $92.0$ & $85.4$ & $82.5$ & $95.7$ & $70.6$ & $91.9$ & $85.9$ & $53.1$ & $69.8$ & $75.3$ \\
            KD-net \cite{pc_klokov2017escape} & $82.3$ & $80.1$ & $74.6$ & $74.3$ & $70.3$ & $88.6$ & $73.5$ & $90.2$ & $87.2$ & $81.0$ & $94.9$ & $57.4$ & $86.7$ & $78.1$ & $51.8$ & $69.9$ & $80.3$ \\
            PN \cite{pc_qi2017pointnet} & $83.7$ & $\mathbf{83.4}$ & $78.7$ & $82.5$ & $74.9$ & $89.6$ & $73.0$ & $91.5$ & $85.9$ & $80.8$ & $95.3$ & $65.2$ & $93.0$ & $81.2$ & $57.9$ & $72.8$ & $80.6$ \\
            SSCNN \cite{Yi_2017_CVPR} & $84.7$ & $81.6$ & $81.7$ & $81.9$ & $75.2$ & $90.2$ & $74.9$ & $\mathbf{93.0}$ & $86.1$ & $\mathbf{84.7}$ & $95.6$ & $66.7$ & $92.7$ & $81.6$ & $60.6$ & $82.9$ & $82.1$ \\
            RSNet \cite{pc_huang2018recurrent}& $84.9$ & $82.7$ & $\mathbf{86.4}$ & $84.1$ & $\mathbf{78.2}$ & $90.4$ & $69.3$ & $91.4$ & $87.0$ & $83.5$ & $95.4$ & $66.0$ & $92.6$ & $81.8$ & $56.1$ & $75.8$ & $82.2$ \\
            PN++ \cite{pc_qi2017pointnet++} & $85.1$ & $82.4$ & $79.0$ & $87.7$ & $77.3$ & $\mathbf{90.8}$ & $71.8$ & $91.0$ & $85.9$ & $83.7$ & $95.3$ & $71.6$ & $94.1$ & $81.3$ & $58.7$ & $76.4$ & $82.6$ \\
            DGCNN \cite{dgcnn} & $85.1$ & $84.2$ & $83.7$ & $84.4$ & $77.1$ & $90.9$ & $\mathbf{78.5}$ & $91.5$ & $87.3$ & $82.9$ & $96.0$ & $67.8$ & $93.3$ & $82.6$ & $59.7$ & $75.5$ & $82.0$ \\
            SGPN \cite{Wang_2018_CVPR} & $\mathbf{85.8}$ & $80.4$ & $78.6$ & $78.8$ & $71.5$ & $88.6$ & $78.0$ & $90.9$ & $83.0$ & $78.8$ & $95.8$ & $\mathbf{77.8}$ & $93.8$ & $87.4$ & $60.1$ & $\mathbf{92.3}$ & $\mathbf{89.4}$ \\
            \hline
            
            \textbf{Morton features} & $81.4$ & $79.9$ & $85.5$ & $\mathbf{92.9}$ & $73.9$ & $81.8$ & $62.7$ & $92.9$ & $\mathbf{90.8}$ & $66.5$ & $\mathbf{98.0}$ & $44.7$ & $\mathbf{95.6}$ & $\mathbf{87.5}$ & $\mathbf{67.9}$ & $74.3$ & $84.7$\\
            \hline
            
            \hline
        \end{tabular}    
    \end{center}
\end{table*}

\vspace{8pt}\noindent\textbf{Experiment (2): Transferring MortonNet to vKITTI.} In this experiment, we use MortonNet pre-trained on S3DIS to extract features on vKITTI point clouds, and use these features in a semantic segmentation pipeline. Similar to the S3DIS case, vKITTI semantic segmentation consist of classifying points into $13$ semantic classes (terrain, tree, vegetation, building, road, guard rail, traffic sign, traffic light, pole, misc, truck, car, and van), however, these are from outdoor scenes and differ significantly form the ones in S3DIS. Although large in the number of points, vKITTI differs from S3DIS in the spatial density of these points. Since vKITTI simulates acquisition from real-world 3D sensors like LiDAR, the data ends up being very sparsely sampled. We show how our S3DIS MortonNet applied to vKITTI can transfer meaningful structure information learned from a densely sampled dataset. In fact, by applying these Morton features to PointNet (in the same fashion as we do for S3DIS), we can beat current state-of-the-art by a significant margin. Experiments in vKITTI are done using the $6$-fold cross-validation splits suggested by \cite{pc_engelmann2018}.

\begin{figure}[!h]
    \centering
    \includegraphics[width=0.99\linewidth]{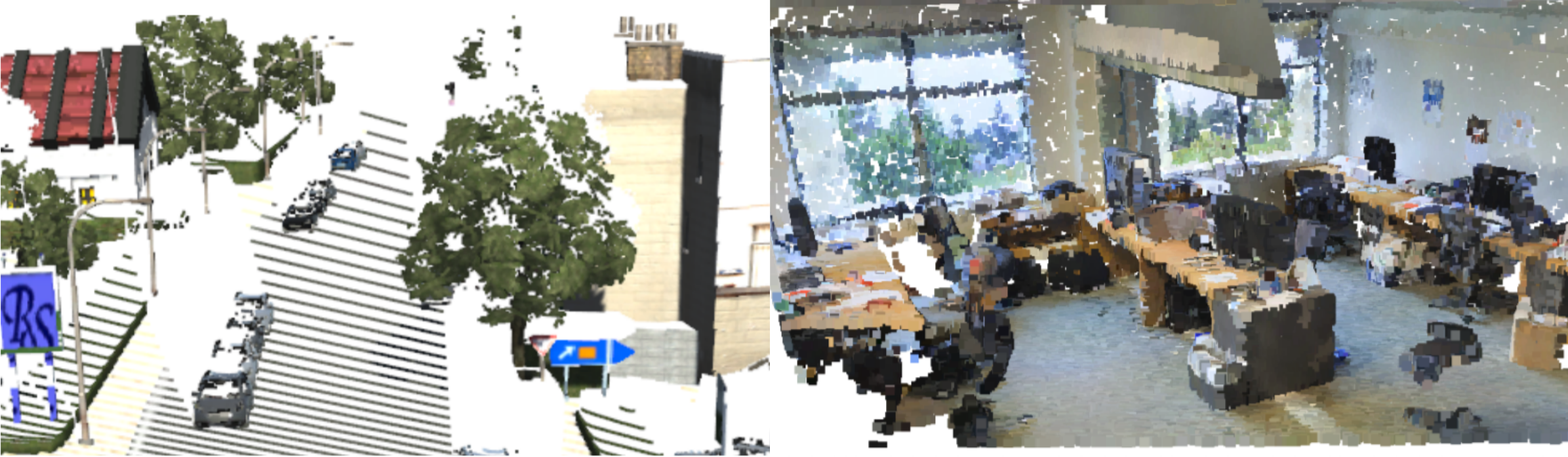}
    \caption{\textbf{Sampling density in vKITTI and S3DIS}. Scenes in vKITTI (left) are sampled much less than in S3DIS (right), since the former simulates acquisition from real-world 3D sensors. 
    }
    \label{fig:vkitti_density}
\end{figure}

\begin{figure}[h]
    \centering
    \includegraphics[width=0.99\linewidth]{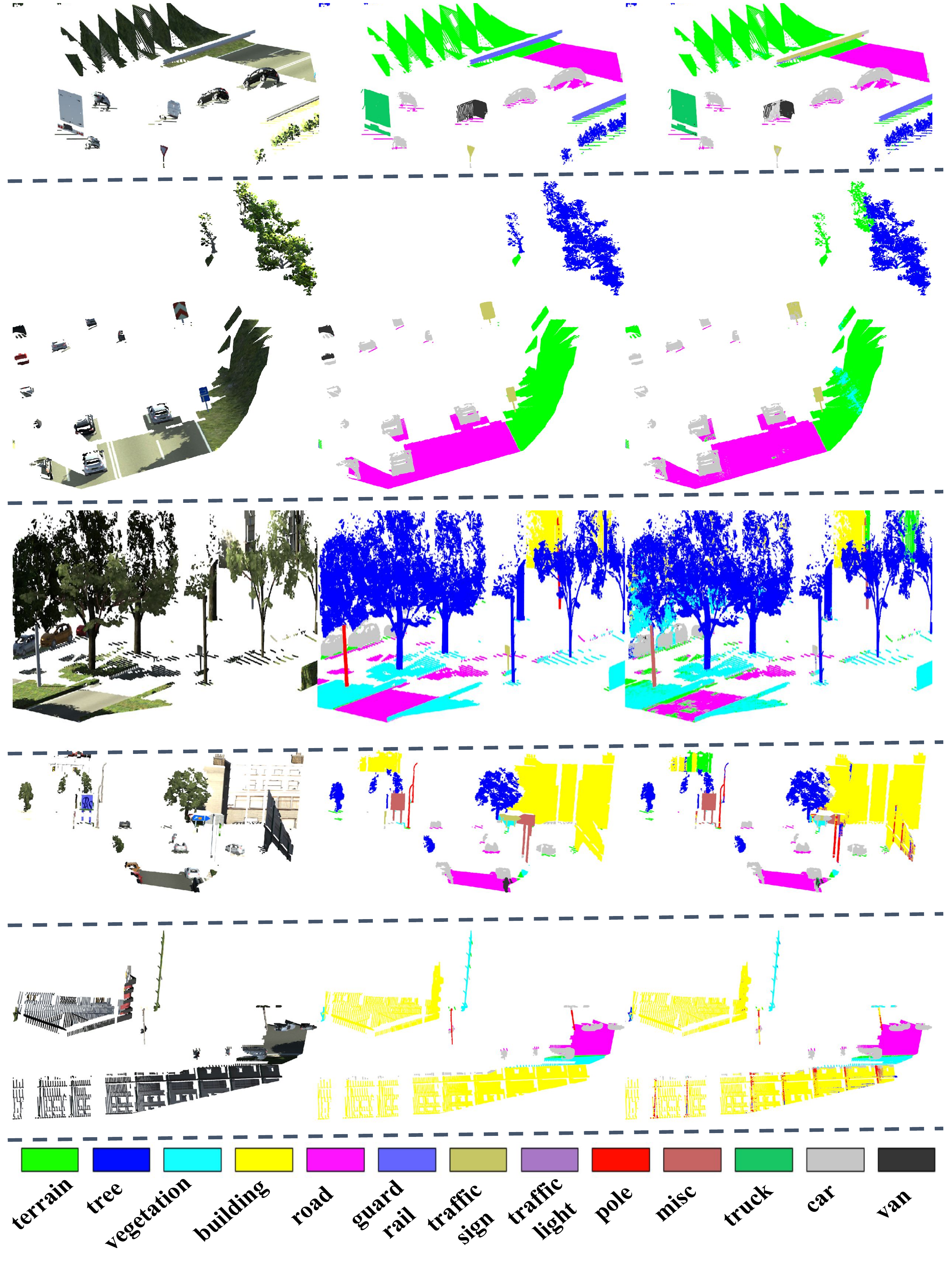}
    \caption{\textbf{Semantic segmentation qualitative results on vKITTI} \cite{pc_engelmann2018}. (\textit{Left}): raw point clouds. (\textit{Middle}): ground truth segmentation. (\textit{Right}): results of PointNet + MortonNet. Better viewed in color and zoomed. Despite the sparse nature of the data, our augmentation manages to properly segment trees, cars, and most roads. Confusion is visible between vegetation and roads.} 
    \label{fig:qualitative_vkitti}
\end{figure}

Table \ref{table:vkitti_seg} shows empirical evidence for a critical strength of MortonNet. By adding Morton features to PointNet (simple model), we improve its performance by $11$ points in mIoU. This makes PointNet + MortonNet the state-of-the-art in semantic segmentation on vKITTI, outperforming the current best and more complex method \cite{pc_ye20183d} by almost $4$ points. We believe one of the reasons why semantic segmentation is challenging in vKITTI is the low sampling density of points in this dataset (refer to Figure \ref{fig:vkitti_density}). During pre-processing, algorithms sample $N$ points from blocks of 1m$\times$1m along the horizontal directions; if a block has less than $N$ points, upsampling is done by randomly repeating points. A consequence of this sampling is that sparse blocks will include less unique points, thus, reducing the amount of context used in training. 
Since Morton features are trained using a very dense dataset (S3DIS), they include local information that is otherwise missed during block based sampling. This information transfers well when computing features on a different, more sparse dataset. Although we do not perform specific ablation studies to prove this fact, we believe this additional information is responsible for the performance increase brought about by incorporating Morton features. 
In Figure \ref{fig:qualitative_vkitti}, we show some  qualitative results of this segmentation task.

\begin{table}[!h]
    \small
    \tabcolsep=0.12cm
    \vspace{6pt}
    \begin{center}
        \caption{\textbf{Part Segmentation on ShapeNet with Varying Training Sizes}. 
        Our approach retains most of its performance even when trained on a very small set of labeled data, while end-to-end approaches severely suffer training data is sparse.} \label{table:sparse_label_training}
        \vspace{8pt}
        \begin{tabular}{l | c c c c c c c }
            \hline
            
            \hline
            
                   & \multicolumn{6}{c}{Training Percentage (\%)} \\
            Method & $5$ & $10$ & $20$ & $40$ & $60$ & $80$ & $100$ \\ 
            
            \hline
            
            \hline
            
            PN \cite{pc_qi2017pointnet} & $59.0$ & $66.4$ & $46.2$ & $76.5$ & $79.0$ & $72.7$ & $\mathbf{83.7}$ \\
            \textbf{MortonNet} & $\mathbf{77.1}$ & $\mathbf{78.0}$ & $\mathbf{78.6}$ & $\mathbf{79.7}$ & $\mathbf{80.0}$ & $\mathbf{80.1}$ & $81.4$ \\
            
            \hline
            
            \hline
        \end{tabular}    
    \end{center}
\end{table}

\vspace{8pt}\noindent\textbf{Experiment (3): Transferring MortonNet to ShapeNet.} Here, we also transfer MortonNet pre-trained on S3DIS to extract features in ShapeNet. The 3D task in ShapeNet is part segmentation, where we need to classify each point in a synthetic CAD model (\eg a plane) by the part it belongs to (\eg body, wing, engine, or tail). Point clouds in ShapeNet are substantially smaller than their S3DIS and vKITTI counterparts, and the underlying task is simpler as evident from its high state-of-the-art performance. 
We therefore show how Morton features can achieve similar results to state-of-the-art methods, but with a much simpler network architecture. We also show our performance is stable when training with much less training data.

Specifically, we compute Morton features on ShapeNet using the S3DIS trained MortonNet. We then pass these features to a simple classifier, comprising only four 1$\times$1 convolution layers interleaved with batch normalization and ReLU layers. Each of the first three convolution layers reduces its input dimensionality by half, while the last one has an output size equal to the number of parts. Refer to the \textbf{supplementary material} for the full details about the classifier architecture. It is evident that such a part segmentation network is much simpler than more complex end-to-end approaches in the literature.  
Table \ref{table:shapenet_seg} summarizes the segmentation results. While our approach does not achieve state-of-the-art average performance on ShapeNet, it surpasses other methods in $7$ out of the $16$ shape categories, especially \emph{rocket} and \emph{cap}. We attribute the large improvements in these classes to the importance of subtle local structure variations between the parts in these classes. Refer to \textbf{supplementary material} for qualitative results.

We show one more set of results with ShapeNet. Since 3D data is expensive to label, it is crucial to develop methods that do not require large labelled datasets. In the spirit of achieving such a goal, we train our simple ShapeNet classifier using Morton Features on a small subset of the part labels in ShapeNet. Table \ref{table:sparse_label_training} shows the effects of decreasing the training percentage for our simple classifier as compared to the end-to-end PointNet \cite{pc_qi2017pointnet} approach. Since our features are trained in a self-supervised manner, our model does not suffer as much as end-to-end models do, when  labeled data in training is limited. In particular, training on $5\%$ of the data only degrades the performance of our simple model by $4.3$ points, while PointNet performance drastically degrades by $24.7$ points.

\section{Conclusion}\label{sec:conc}
We presented a self-supervised approach to learning local point-wise features from unstructured 3D point clouds. We are the first to apply such approach to point cloud data. We showed the importance of learning from Z-ordered (Morton-ordered) sequences, and demonstrated the effectiveness of Morton features for point-based 3D tasks. Although we trained our self-supervised model with only one dataset, we showed that our features can generalize well across datasets. We showed significant improvement when augmenting other methods with our features. We also showed how our features can help with more stable training in the presence of smaller training sets. Although we only used them for segmentation, Morton features are generic and effective in encoding local point structure, and we believe they could also help enhance other 3D tasks.

{\small
\bibliographystyle{ieee}
\bibliography{egbib}
}

\end{document}